\def\year{2017}\relax
\documentclass[letterpaper]{article}
\usepackage{aaai17}
\usepackage{times}
\usepackage{helvet}
\usepackage{courier}
\usepackage{url}
\usepackage{latexsym}
\usepackage{graphicx}
\usepackage{pgfplots}
\usepackage{amsmath}
\usepackage{amssymb}
\usepackage{mathrsfs}
\usepackage{caption}
\usepackage{array}
\newcolumntype{P}[1]{>{\raggedright\arraybackslash}p{#1}}
\newcommand{\specialcell}[2][c]{%
  \begin{tabular}[#1]{@{}l@{}}#2\end{tabular}}
\usetikzlibrary{patterns}

\newcommand\Tstrut{\rule{0pt}{2.0ex}}       

\let\OLDthebibliography\thebibliography
\renewcommand\thebibliography[1]{
  \OLDthebibliography{#1}
  \setlength{\parskip}{0pt}
  \setlength{\itemsep}{0pt plus 0.3ex}
}

\title{\#SarcasmDetection is soooo general!\\Towards a Domain-Independent Approach for Detecting Sarcasm}
\author{Natalie Parde \and Rodney D. Nielsen\\
Department of Computer Science and Engineering\\
University of North Texas\\
\url{{natalie.parde, rodney.nielsen}@unt.edu}}
	
\begin{document}

\maketitle

\begin{abstract}
Automatic sarcasm detection methods have traditionally been designed for maximum performance on a specific domain.  This poses challenges for those wishing to transfer those approaches to other existing or novel domains, which may be typified by very different language characteristics.  We develop a general set of features and evaluate it under different training scenarios utilizing in-domain and/or out-of-domain training data.  The best-performing scenario, training on both while employing a domain adaptation step, achieves an F\textsubscript{1} of 0.780, which is well above baseline F\textsubscript{1}-measures of 0.515 and 0.345.  We also show that the approach outperforms the best results from prior work on the same target domain.
\end{abstract}

\section{Introduction}

Sarcasm, a creative device used to communicate an intended meaning that is actually the opposite of its literal meaning,\footnote{\url{merriam-webster.com/dictionary/sarcasm}} is notoriously difficult to convey and interpret through text, in part because doing so relies heavily upon shared contextual understandings that can be marked more easily by altered prosody (e.g., emphasis upon certain words) or non-verbal signals (e.g., rolling one's eyes).  It is a complex process even for humans, and in fact an inability to detect sarcasm has been linked with a number of neurocognitive disorders, including dementia \cite{kipps2009understanding}.  It is similarly a challenging open task in natural language processing, and has direct implications to a number of other critical application areas, such as sentiment analysis.

Most research on automatic sarcasm detection to date has focused on the Twitter domain, which boasts an ample source of publicly-available data, some of which is already self-labeled by users for the presence of sarcasm (e.g., with \#sarcasm).  However, Twitter is highly informal, space-restricted, and subject to frequent topic fluctuations from one post to the next due to the ebb and flow of current events---in short, it is not broadly representative of most text domains.  Thus, sarcasm detectors trained using features designed for maximum Twitter performance are not necessarily transferable to other domains.  Despite this, it is desirable to develop approaches that can harness the more generalizable information present in the abundance of Twitter data.

In this work, we develop a set of domain-independent features for sarcasm detection and show that the features generally perform well across text domains.  Further, we validate that domain adaptation can be applied to sarcasm detection to leverage patterns in out-of-domain training data, even when results from training only on that source domain data are extremely bad (far below baseline results), to improve over training on only the target data or over training on the simply combined dataset.  Finally, we make a new dataset of sarcastic and non-sarcastic tweets available online as a resource to other researchers.\footnote{\url{hilt.cse.unt.edu/resources.html}}

\section{Related Work}
The majority of work on automatic sarcasm detection has been done using Twitter, and to a smaller extent Amazon product reviews.  Research outside of those domains has been scarce, but interesting.  Notably, Burfoot and Baldwin \shortcite{Burfoot:2009:ASD:1667583.1667633} automatically detected satirical news articles using unigrams, lexical features, and semantic validity features, and Justo et al. \shortcite{Justo2014124} used n-gram, linguistic, and semantic features to detect the presence of sarcasm in the Internet Argument Corpus \cite{walker2012corpus}.  The remainder of this section describes prior work with Twitter and Amazon.

\subsection{Sarcasm Detection on Twitter}
\label{related_work_twitter}

Twitter is a micro-blogging service that allows users to post short ``tweets'' to share content or describe their feelings or opinions in 140 characters or less.  For researchers, it boasts a low cost of annotation and plentiful supply of data (users often self-label their tweets using the ``\#'' symbol---many explicitly label their sarcastic tweets using the hashtag ``\#sarcasm'').  A variety of approaches have been taken toward automatically detecting sarcasm on Twitter, including explicitly using the information present in a tweet's hashtag(s); Maynard and Greenwood \shortcite{maynard2014cares} learned which hashtags characteristically corresponded with sarcastic tweets, and used the presence of those indicators to predict other sarcastic tweets, with high success.  \citeauthor{liebrecht2013perfect} \shortcite{liebrecht2013perfect} detected sarcasm in Dutch tweets using unigram, bigram, and trigram features.

\citeauthor{Rajadesingan:2015:SDT:2684822.2685316} \shortcite{Rajadesingan:2015:SDT:2684822.2685316} detected sarcastic tweets based on features adapted from behavioral models of sarcasm usage, drawing extensively from individual users' Twitter histories and relying heavily on situational context and user characteristics.  The system also employed lexical features and grammatical correctness as a means of modelling different aspects of the user's behavior.

Other researchers have had success identifying sarcasm by a tweet's use of positive sentiment to describe a negative situation \cite{RiloffSarcasm}, employing contextual \cite{BammanICWSM1510538} or pragmatic \cite{Gonzalez-Ibanez:2011:IST:2002736.2002850} features, and observing the writing style and emotional scenario of a tweet \cite{Reyes2013}.  An underlying theme among these methods is that the features are generally designed specifically for use with tweets.  A major challenge in developing a more general approach for sarcasm detection lies in developing features that are present across many domains, yet still specific enough to reliably capture the differences between sarcastic and non-sarcastic text.

Finally, some researchers have recently explored approaches that rely on word embeddings and/or carefully tailored neural networks, rather than on task-specific feature design \cite{ghosh-guo-muresan:2015:EMNLP,ghosh-veale:2016:WASSA2016,AmirCoNLL}.  Since neural networks offer little transparency, it is uncertain whether the features learned in these approaches would be easily transferable across text domains for this task (prior research on other tasks suggests that the features computed by deep neural networks grow increasingly specific to the training dataset---and in turn, to the training domain---with each layer \cite{Yosinski:2014:TFD:2969033.2969197}).  Although an interesting question, the focus herein is on uncovering the specific types of features capable of leveraging general patterns for sarcasm detection, and this can be more easily examined using shallower learning algorithms.

\subsection{Sarcasm Detection on Amazon Reviews}
\label{related_work_amazon}
Research on automatic sarcasm detection in other domains has been limited, but recently a publicly-available corpus of sarcastic and non-sarcastic Amazon product reviews was released by Filatova \shortcite{FILATOVA12.661} to facilitate research.  \citeauthor{buschmeier-cimiano-klinger:2014:W14-26} \shortcite{buschmeier-cimiano-klinger:2014:W14-26} test many feature combinations on this dataset, including those based on metadata (e.g., Amazon star rating), sentiment, grammar, the presence of interjections (e.g., ``wow'') or laughter (e.g., through onomatopoeia or acronyms such as ``lol''), the presence of emoticons, and bag-of-words features.  Their highest F\textsubscript{1} (0.744) is achieved using all of these with a logistic regression classifier; however, using only the star rating, they still achieve an F\textsubscript{1} of 0.717.  This highlights the need for high-performing, general features for sarcasm detection; metadata features are highly domain-specific, and even bag-of-words trends may be unique to certain domains (``trump'' was one of the most common unigrams in our own Twitter training set, but only occurred once across all Amazon product reviews).

Prior to the release of Filatova's dataset, \citeauthor{davidov-tsur-rappoport:2010:CONLL} \shortcite{davidov-tsur-rappoport:2010:CONLL} developed a semi-supervised approach to classify tweets or Amazon reviews as sarcastic or non-sarcastic by clustering samples based on grammatical features and the full or partial presence of automatically-extracted text patterns.  They evaluated their work on a sample of the classified instances annotated by anonymous users on Amazon Mechanical Turk.  They tested several different seed sets with their approach, one of which contained a mixture of positive Amazon reviews, positive \#sarcasm-tagged tweets, and a manually-selected sample of negative tweets.  Although they did not report test results on Amazon reviews using this seed set, they did report test results on \#sarcasm-tagged tweets, achieving an F-measure of 0.545.  Their work is the closest to ours, because it attempts to harness training samples from both the Twitter and Amazon review domains.

\section{Methods}
\subsection{Data Collection}

\setlength{\textfloatsep}{10pt plus 1.0pt minus 2.0pt}
\begin{table}[t]
\centering
\footnotesize
\begin{tabular}{p{1.2cm}|p{2.9cm}|p{2.9cm}}
        & \textbf{Sarcastic (Train/Test)} & \textbf{Non-Sar. (Train/Test)} \\ \hline
Twitter & 1959 (1568/391)     & 3039 (2430/609) \Tstrut\\ \hline
Amazon  & 437  (350/87)       & 817 (653/164)   \Tstrut\\ \hline
\end{tabular}
\caption{Twitter and Amazon Dataset Distributions}
\label{data_info}
\end{table}

Data was taken from two domains: Twitter, and Amazon product reviews.  The Amazon reviews were from the publicly available sarcasm corpus developed by Filatova \shortcite{FILATOVA12.661}.  To build our Twitter dataset, tweets containing exactly one of the trailing hashtags ``\#sarcasm,'' ``\#happiness,'' ``\#sadness,'' ``\#anger,'' ``\#surprise,'' ``\#fear,'' and ``\#disgust'' were downloaded regularly during February and March 2016.  Tweets containing the latter six hashtags, corresponding to Ekman's six basic emotions \cite{Ekman1992}, were labeled as non-sarcastic.  Those hashtags were chosen because their associated tweets were expected to still express opinions, similarly to sarcastic tweets, but in a non-sarcastic way.  Tweets containing \#sarcasm were labeled as sarcastic; annotating tweets with the \#sarcasm hashtag as such is consistent with the vast majority of prior work in the Twitter domain \cite{Gonzalez-Ibanez:2011:IST:2002736.2002850,liebrecht2013perfect,maynard2014cares,Rajadesingan:2015:SDT:2684822.2685316,BammanICWSM1510538,ghosh-guo-muresan:2015:EMNLP,AmirCoNLL}.

The downloaded tweets were filtered to remove retweets, ``@replies,'' and tweets containing links.  Retweets were removed to avoid having duplicate copies of identical tweets in the dataset, @replies were removed in case the hashtag referred to content in the tweet to which it replied rather than content in the tweet itself, and tweets with links were likewise removed in case the hashtag referred to content in the link rather than in the tweet itself.  Requiring that the specified hashtag trailed the rest of the tweet (it could only be followed by other hashtags) was done based on the observation that when sarcastic or emotional hashtags occur in the main tweet body, the tweet generally discusses sarcasm or the specified emotion, rather than actually expressing sarcasm or the specified emotion.  Finally, requiring that only one of the specified hashtags trailed the tweet eliminated cases of ambiguity between sarcastic and non-sarcastic tweets.  All trailing ``\#sarcasm'' or emotion hashtags were removed from the data before training and testing, and both datasets were randomly divided into training (80\%) and testing (20\%) sets.  Further details are shown in Table \ref{data_info}.

\subsection{Features}
Three feature sets were developed (one general, and two targeted toward Twitter and Amazon, respectively).  Resources used to develop the features are described in Table \ref{lexical_resources}.  Five classifiers (Na\"{\i}ve Bayes, J48, Bagging, DecisionTable, and SVM), all from the Weka\footnote{\url{cs.waikato.ac.nz/ml/weka/}} library, were tested using five-fold cross-validation on the training sets, and the highest-scoring (Na\"{\i}ve Bayes) was selected for use on the test set.

\begin{table}
\centering
\footnotesize
\begin{tabular}{p{1.25cm}|p{6.3cm}}
\textbf{Resource}   & \textbf{Description} \\ \hline
Liu05               & Opinion lexicon containing 2006 pos. words and 4783 neg. words \cite{Liu:2005:OOA:1060745.1060797}.    \\ \hline
MPQA                & Subjectivity lexicon containing strongly or weakly subjective positive (2718) or negative (4910) words \cite{Wilson:2005:RCP:1220575.1220619}.  \\ \hline
AFINN               & Sentiment lexicon for microblogs \cite{Hansen2011} containing 2477 words/phrases labeled with values from -5 (negative) to +5 (positive).    \\ \hline
Google Web1T & Large collection of n-grams and their frequencies scraped from the web \cite{GoogleWeb1T}.             \\ \hline       
\end{tabular}
\caption{Lexical Resources}
\label{lexical_resources}
\end{table}

\subsubsection{Domain-Specific Features}
The Twitter- (\textit{T}) and Amazon-specific (\textit{A}) features are shown in Table \ref{twitter_features}.  Domain-specific features were still computed for instances from the other domain unless it was impossible to compute those features in that domain (i.e., Amazon Star Rating for Twitter instances), in which case they were left empty.  Twitter-specific features are based on the work of \citeauthor{maynard2014cares} \shortcite{maynard2014cares} and \citeauthor{RiloffSarcasm} \shortcite{RiloffSarcasm}.  Maynard and Greenwood detect sarcastic tweets by checking for the presence of learned hashtags that correspond with sarcastic tweets, as well as sarcasm-indicator phrases and emoticons.  We construct binary features based on their work, and on Riloff et al.'s work \shortcite{RiloffSarcasm}, which determined whether or not a tweet was sarcastic by checking for positive sentiment phrases contrasting with negative situations (both of which were learned from other sarcastic tweets).  We also add a feature indicating the presence of laughter terms.  Amazon-based features are primarily borrowed from \citeauthor{buschmeier-cimiano-klinger:2014:W14-26}'s \shortcite{buschmeier-cimiano-klinger:2014:W14-26} earlier work on the Amazon dataset.
\footnotetext[4]{Individual binary features for each of the sarcasm hashtags (5 features) and laughter tokens (9 features) were also included.}

\begin{table}[t]
\centering
\footnotesize
\begin{tabular}{P{1.83cm}|p{5.74cm}}
\textbf{Feature}   & \textbf{Description} \\ \hline
\specialcell{Contains Sar-\\casm Hashtag}  & \specialcell{\textit{(T)} True if contains hashtag learned by May-\\nard \& Greenwood (excluding \#sarcasm).\footnotemark}    \\ \hline
\specialcell{Contains Sar-\\castic Smiley} & \specialcell{\textit{(T)} True if instance contains a sarcastic emo-\\ticon learned by Maynard \& Greenwood.}  \\ \hline
Contains Sar. Indicator & \textit{(T)} True if instance contains a sarcasm indicator phrase learned by Maynard \& Greenwood.    \\ \hline
\specialcell{Contains Pos-\\itive Predicate} & \specialcell{\textit{(T)} True if instance contains a positive \\predicate learned from Twitter by Riloff.}             \\ \hline   
Contains Pos. Sentiment & \textit{(T)} True if instance contains a positive sentiment phrase learned from Twitter by Riloff. \\ \hline   
Contains Neg. Situation & \textit{(T)} True if instance contains a negative situation phrase learned from Twitter by Riloff. \\ \hline
Pos. Sent. Precedes Neg. Situation & \textit{(T)} True if contains a pos. predicate or sentiment phrase learned by Riloff that precedes a learned neg. situation phrase by $\leq$ 5 tokens. \\ \hline
Contains Laughter & \textit{(T)} True if instance contains \textit{hahaha}, \textit{haha}, \textit{hehehe}, \textit{hehe}, \textit{jajaja}, \textit{jaja}, \textit{lol}, \textit{lmao}, or \textit{rofl}.\footnotemark[\value{footnote}] \\ \hline
Star \mbox{Rating}  & \textit{(A)} Numeric score (1-5) corresponding to number of stars associated with the review. \\ \hline
Contains \textit{Wow} & \textit{(A)} True if the instance contains \textit{wow}.  \\ \hline
Contains \textit{Ugh} & \textit{(A)} True if the instance contains \textit{ugh}.    \\ \hline
Contains \textit{Huh} & \textit{(A)} True if the instance contains \textit{huh}. \\ \hline   
Contains ``...'' & \textit{(A)} True if the instance contains an ellipsis. \\ \hline
\end{tabular}
\caption{Domain-Specific Features}
\label{twitter_features}
\end{table}

\subsubsection{General Features}
\label{General_Features}
We model some of our general features after those from \citeauthor{RiloffSarcasm} \shortcite{RiloffSarcasm}, under the premise that the underlying principle that sarcasm often associates positive expressions with negative situations holds true across domains.  Since positive sentiment phrases and negative situations learned from tweets are unlikely to generalize to different domains, we instead use three sentiment lexicons to build features that capture positive and negative sentiment rather than checking for specific learned phrases.  Likewise, rather than bootstrapping specific negative situations from Twitter, we calculate the pointwise mutual information (PMI) between the most positive or negative word in the instance and the n-grams that immediately proceed it\footnote{Frequencies for computing PMI were from Google Web1T.} to create a more general version of the feature.  Other general features developed for this work rely on syntactic characteristics, or are bag-of-words-style features corresponding to the tokens most strongly correlated or most common in sarcastic and non-sarcastic instances from Twitter and Amazon training data.  All general features are outlined in Table \ref{table:general_features}.

\setlength{\abovedisplayskip}{3pt}
\setlength{\belowdisplayskip}{3pt}
\begin{table*}[!ht]
\centering
\footnotesize
\begin{tabular}{p{3.1cm}|p{13.1cm}}
\textbf{Feature}  & \textbf{Description} \\ \hline
Most Polar Unigram                                                   & The single most positive or negative unigram in the instance, according to AFINN. \\ \hline
Most Polar Score                                                     & The score, ranging from -5 to +5, corresponding to the most polar unigram.\\ \hline
\specialcell{\% Strongly (Weakly) \\Subj. Pos. (Neg.) Words} & \specialcell{Four features: The number of words identified as strongly (weakly) subjective positive (negative) words \\in the instance according to MPQA, divided by the count of the instance's words found in MPQA.} \\ \hline
Avg. Polarity of the \newline Instance (Liu05)                             & The sum of the polarity scores for all words (positive words = $+1$, negative words = $-1$) in the instance included in Liu05, divided by the total number of words in the instance included in Liu05.  \\ \hline
Avg. Polarity of the \newline Instance (MPQA)                              & An analogue of the above, for MPQA.  Strongly subjective pos. words are assigned a score of +2, weakly subjective pos. words +1, weakly subjective neg. words -1, and strongly subjective neg. words -2.\footnotemark   \\ \hline
Avg. Polarity of the \newline Instance (AFINN)                             & The sum of the polarity scores for all words in the instance included in the AFINN sentiment lexicon, divided by the total number of words in the instance included in that lexicon.\\ \hline
Overall Polarity of \newline the Instance                                     & Three separate scores, corresponding to the sum of the polarity scores for all words in the Liu05, MPQA, and AFINN lexicons, respectively (scores are calculated as described above). \\ \hline
\specialcell{\% Positive (Negative) \\Words}                              & \specialcell{Six separate features, calculated by dividing the number of positive (or negative) words in the instance \\according to Liu05, MPQA, and AFINN, respectively, by the total number of words in the instance.} \\ \hline
N-gram PMI Scores                                                    & Four features corresponding to the PMI between the most polar unigram and the 1-, 2-, 3-, and 4-grams that immediately follow it.  Let $\mathrm{p}(w,W_n)$ be the probability of the sequence starting with unigram $w$ and ending with the n-gram $W_n$ of size $n$, where $\mathrm{C}(w,W_n)$ is the number of occurrences of $w$ immediately followed by $W_n$, and $N$ is the count of all n-grams of length $n+1$ in the corpus.  Then,
\begin{equation}
\mathrm{p}(w,W_n) = \frac{1}{N}\mathrm{C}(w,W_n), \qquad \mathrm{PMI}(w,W_n) = \log\frac{\mathrm{p}(w,W_n)}{\mathrm{p}(w,*_n) \times \mathrm{p}(*,W_n)}
\end{equation}
where $*_n$ can be any n-gram of length $n$ and $*$ can be any unigram.  In tweets, hashtags are removed prior to calculating PMI (e.g., ``\#love'' becomes ``love''), and any tokens beginning with ``@'' may be matched by any token (these are assumed to be mentions of another Twitter user by his or her username). \\ \hline
All-Caps Words                              & Two features, corresponding to the raw number of all-caps words in the instance, and the number of all-caps words divided by the total number of words in the instance. \\ \hline
Consecutive Chars.                                     & The highest number of consecutive repeated characters in the instance (e.g., ``Sooooo'' $\Rightarrow 5$). \\ \hline
Consecutive Punct.                              & The highest number of consecutive punctuation characters in the instance.\\ \hline
\specialcell{Specific Character \\Features}                                          & \specialcell{Two binary features: one is equal to 1 if the instance contains an exclamation mark, and the other is \\equal to 1 if the instance contains a question mark.} \\ \hline
Largest Score Gap                                                    & The most negative score in the instance, according to the AFINN lexicon, subtracted from the most positive score in the instance, according to the AFINN lexicon.  \\ \hline
Bag-of-Associated-Words (BOAW)                         & Up to 200 features: training instances were divided into four groups (Sarcastic $\times$ Non-Sarcastic) $\times$ (Amazon $\times$ Twitter).  For each group, the 50 unigrams most strongly correlated with that class and domain were computed based on the PMI between the unigram and class label.  Specifically: $\mathrm{PMI}(w,l) = \log\frac{\mathrm{p}(w,l)}{\mathrm{p}(w) \times \mathrm{p}(l)}$, where $w$ is the unigram, $l$ is a label, $\mathrm{p}(w,l)$ is the joint probability of an instance containing $w$ and being labeled $l$, $\mathrm{p}(w)$ is the probability of $w$ being in any instance, and $\mathrm{p}(l)$ is the fraction of instances labeled $l$.  Probabilities were computed separately for Amazon and Twitter; stopwords are removed prior to calculating PMI.  Plus-one smoothing was used for all probabilities.
\\ \hline
Bag-of-Common-Words (BOCW)                             & Up to 200 features: training instances were divided into the same four groups as above.  For each group the 50 most common unigrams were determined.  Any duplicates across groups were then removed.  \\ \hline
\end{tabular}
\caption{General Feature Set}
\label{table:general_features}
\end{table*}

\section{Evaluation}
The features used for each train/test scenario are shown in the first column of Table \ref{full_analysis}.  \textit{Twitter Features} refers to all features listed in Table \ref{twitter_features} preceded by the parenthetical \textit{(T)}, and \textit{Amazon Features} to all features preceded by \textit{(A)}.  \textit{General: Other Polarity} includes the positive and negative percentages, average polarities, overall polarities, and largest polarity gap features from Table \ref{table:general_features}.  \textit{General: Subjectivity} includes the \% strongly subjective positive words, \% weakly subjective positive words, and their negative counterparts.  We also include two baselines: the \textit{All Sarcasm} case assumes that every instance is sarcastic, and the \textit{Random} case randomly assigns each instance as sarcastic or non-sarcastic.

Results are reported for models trained only on Twitter, only on Amazon, on both training sets, and on both training sets when Daum\'e's \shortcite{daumeiii:2007:ACLMain} EasyAdapt technique is applied, employing Twitter as the algorithm's source domain and Amazon as its target domain.  
EasyAdapt works by modifying the feature space so that it contains three mappings of the original features: a general (source + target) version, a source-only version, and a target-only version.  More specifically, assuming an input feature set $\mathcal{X} = \mathbb{R}^F$ for some $F > 0$, where $F$ is the number of features in the set, EasyAdapt transforms $\mathcal{X}$ to the augmented set, $\mathcal{\check{X}} = \mathbb{R}^{3F}$.  
The mappings $\Phi^s, \Phi^{t} : \mathcal{X} \rightarrow \mathcal{\check{X}}$ for the source and target domain data, respectively, are defined as $\Phi^s(\mathbf{x}) = \langle\mathbf{x},\mathbf{x},\mathbf{0}\rangle$ and $\Phi^t(\mathbf{x}) = \langle\mathbf{x},\mathbf{0},\mathbf{x}\rangle$, where $\mathbf{0} = \langle 0,...,0 \rangle \in \mathbb{R}^F$ is the zero vector.  Refer to Daum\'e \shortcite{daumeiii:2007:ACLMain} for an in-depth discussion of this technique.

Each model was tested on the Amazon test data (the model trained only on Twitter was also tested on the Twitter test set).  Amazon reviews were selected as the target domain since the Twitter dataset was much larger than the Amazon dataset; this scenario is more consistent with the typically stated goal of domain adaptation (a large labeled out-of-domain source dataset and a small amount of labeled data in the target domain), and most clearly highlights the need for a domain-general approach.  
\footnotetext[6]{Part-of-speech is considered in MPQA; Amazon and Twitter data was tagged using Stanford CoreNLP \cite{manning-EtAl:2014:P14-5} and the Twitter POS-tagger \cite{Owoputi13improvedpart-of-speech}, respectively.} 

Finally, we include the best results reported by \citeauthor{buschmeier-cimiano-klinger:2014:W14-26}  \shortcite{buschmeier-cimiano-klinger:2014:W14-26} on the same Amazon dataset.  For a more direct comparison between our work and theirs, we also report the results from using all of our features under the same classification conditions as theirs (10-fold cross-validation using \textit{scikit-learn}'s Logistic Regression,\footnote{\url{scikit-learn.org}} tuning with an F\textsubscript{1} objective).  We refer to the latter case as \textit{Our Results, Same Classifier as Prior Best}.

\subsection{Results}
\begin{table*}[t]
\centering
\footnotesize
\begin{tabular}{p{3cm}|ccc|ccc|ccc|ccc||ccc}
\hline
& \multicolumn{12}{c||}{\textbf{Test on Amazon Reviews}} & \multicolumn{3}{c}{\textbf{Test on Twitter}} \\
\hline
                                         & \multicolumn{3}{c}{\textbf{Train on Twitter}} & \multicolumn{3}{c}{\textbf{Train on Both}} & \multicolumn{3}{c}{\textbf{Train on Amazon}} & \multicolumn{3}{c||}{\textbf{EasyAdapt}} & \multicolumn{3}{c}{\textbf{Train on Twitter}}\\ \hline
                                         & \textbf{P}          & \textbf{R}          & $\mathbf{F_{1}}$          & \textbf{P}         & \textbf{R}         & $\mathbf{F_{1}}$         & \textbf{P}          & \textbf{R}          & $\mathbf{F_{1}}$         & \textbf{P}     & \textbf{R}     & $\mathbf{F_{1}}$     & \textbf{P} & \textbf{R} & $\mathbf{F_{1}}$        \\ \hline
Baseline: All Sarcasm                                 & .35       & 1.0        & .515       & .35      & 1.0       & .515      & .35       & 1.0        & .515      & .35  & 1.0   & .515          & .39 & 1.0 & .562 \\ \hline
Baseline: Random                                 & .35       & .35        & .347       & .35      & .35       & .347      & .35       & .35        & .347      & .35  & .35   & .347          & .39 & .39 & .391 \\ \hline\hline
Twitter Features                & .32       & .36       & .337       & .00       & .00       & .000       & .00        & .00        & .000       & .00   & .00   & .000        & .57 & .24 & .341    \\ \hline
Amazon Features                 & .48        & .26        & .341        & .76      & .68      & .715      & .76       & .67       & .712      & .76  & .68  & .715     & .47 & .14 & .216      \\ \hline
Gen.: Most Polar Word & .26       & .30       & .275       & .43      & .15      & .222      & .68       & .20       & .304      & .52  & .33  & .406         & .65 & .38 & .479 \\ \hline
Gen.: Most Polar Score & .26       & .48       & .335       & .00     & .00     & .000     & .59       & .46       & .516      & .59  & .46  & .516         & .56 & .40 & .466 \\ \hline
General: Other Polarity   & .26       & .36       & .302       & .35      & .40      & .374      & .61       & .69       & .645      & .55  & .67  & .601         & .51 & .58 & .542 \\ \hline
General: Subjectivity  & .25        & .02       & .042       & .25       & .02      & .042      & .60       & .36       & .446      & .59  & .36  & .443   & .48 & .15 & .229       \\ \hline
General: Syntactic  & .51       & .26       & .348       & .69      & .36      & .470      & .71       & .43       & .532      & .65  & .51  & .568         & .46 & .17 & .250 \\ \hline
General: PMI Features                    & .00        & .00        & .000        & .25      & .03      & .061      & .34       & .14       & .197      & .42  & .21  & .277     & .60 & .02 & .030     \\ \hline
General: BOAW & .00      & .00      & .000      & .00     & .00     & .000     & .63       & .20       & .298      & .60  & .23  & .331    & .00 & .00 & .000      \\ \hline
General: BOCW & .47       & .59       & .523       & .58      & .16      & .252       & .59       & .60       & .594      & .63  & .46 & .530         & .59 & .39 & .467 \\ \hline
All General Features                     & .26       & .32       & .290        & .42      & .29      & .340     & .63       & .69       & .659      & .69  & .67  & .678      & .55 & .62 & .582    \\ \hline
All Features                             & .25       & .31       & .276       & .66      & .54       & .595      & .66       & .77       & .713      & .75  & .82  & \textbf{.780} & .55 & .62 & .583\\ \hline \hline
\multicolumn{7}{l|}{Prior Best Results \cite{buschmeier-cimiano-klinger:2014:W14-26}} & .82 & .69 & .744 & \multicolumn{6}{l}{} \\ \hline
\multicolumn{7}{l|}{Our Results, Same Classifier as Prior Best} & .80 & .71 & \textbf{.752} & \multicolumn{6}{l}{} \\ \hline
\end{tabular}
\caption{Test Results --- Full Analysis}
\label{full_analysis}
\end{table*}

The results, including each of the training scenarios noted earlier, are presented in Table \ref{full_analysis}.  Precision, recall, and F\textsubscript{1} on the positive (sarcastic) class were recorded.
The highest F\textsubscript{1} achieved (0.780) among all cases was from training on the EasyAdapted Twitter and Amazon data.  In comparison, training only on the Amazon reviews produced an F\textsubscript{1} of 0.713 (training and testing only on Amazon reviews with our features but with the same classifier and cross-validation settings as \citeauthor{buschmeier-cimiano-klinger:2014:W14-26} \shortcite{buschmeier-cimiano-klinger:2014:W14-26} led to an F\textsubscript{1} of 0.752, outperforming prior best results on that dataset).  Training on both without EasyAdapt led to an F\textsubscript{1} of 0.595 (or 0.715 when training only on Amazon-specific features), and finally, training only on Twitter data led to an F\textsubscript{1} of 0.276.  Training and testing on Twitter produced an F\textsubscript{1} of 0.583 when training on all features.\footnote{Further analysis of the Twitter-specific features revealed that contains\_sarcasm\_hashtag, contains\_sarcastic\_smiley, and contains\_sarcasm\_indicator\_phrase all led to F\textsubscript{1}$=0.0$ when used individually; although these performed quite well in prior work, our Twitter dataset did not contain the indicators with high enough frequency to have any impact on the overall classification outcome.}

\section{Discussion}

When testing on Amazon reviews, the worst-performing case was that in which the classifier was trained only on Twitter data (it did not manage to outperform either baseline).  This underscores the inherent variations in the data across the two domains; despite the fact that many of the features were deliberately designed to be generalizable and robust to domain-specific idiosyncrasies, the different trends across domains still confused the classifier.  

However, combining all of that same Twitter data with a much smaller amount of Amazon data (3998 Twitter training instances relative to 1003 Amazon training instances) and applying EasyAdapt to the combined dataset performed quite well ($F_{1}$=0.780).  The classifier was able to take advantage of a wealth of additional Twitter samples that had led to terrible performance on their own ($F_{1}$=0.276).  Thus, the high performance demonstrated when the EasyAdapt algorithm is applied to the training data from the two domains is particularly impressive.  It shows that more data is indeed better data---provided that the proper features are selected and the classifier is properly guided in handling it.

Overall, the system cut the error rate from .256 to .220, representing a 14\% relative reduction in error over prior best results on the Amazon dataset.  Our results testing on Twitter are not directly comparable to others, since prior work's datasets could not be released; however, our results ($F_{1}$=0.583) are in line with those reported previously (\citeauthor{RiloffSarcasm} \shortcite{RiloffSarcasm}: $F_{1}$=0.51; \citeauthor{davidov-tsur-rappoport:2010:CONLL} \shortcite{davidov-tsur-rappoport:2010:CONLL}: $F_{1}$=0.545).  Additionally, our Twitter data did not contain many indicators shown to be discriminative in the past (leading our general features to be better predictors of sarcasm even when training/testing entirely within the domain), and our focus in developing features was on general performance rather than performance on Twitter specifically.

\begin{figure}
	\centering
		\includegraphics[width=8cm]{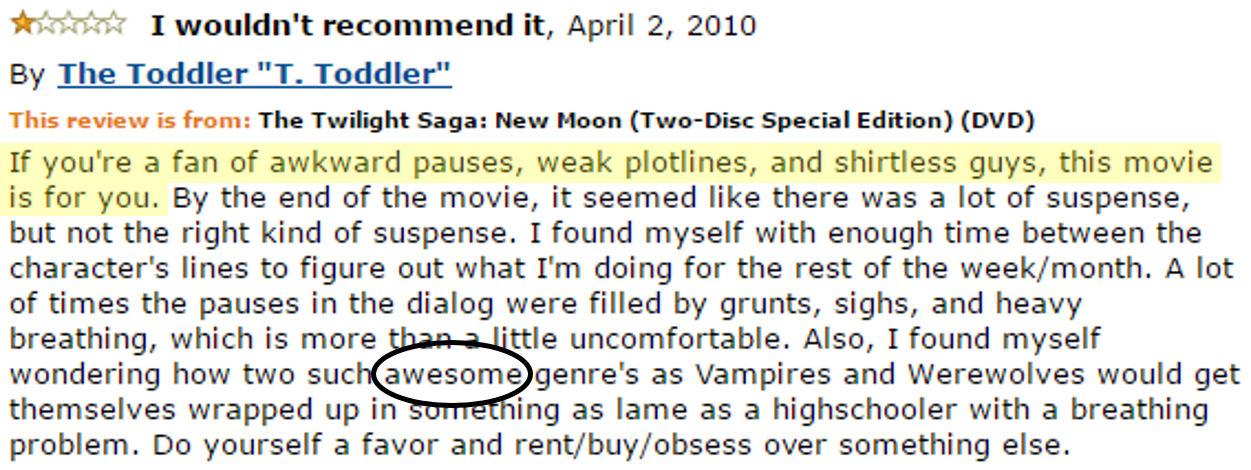}
	\caption{Example from Amazon Product Reviews}
	\label{fig:Snippet1}
\end{figure}

Both datasets were somewhat noisy.  Many full-length reviews that were marked as ``sarcastic'' were only partially so, and included other sentences that were not sarcastic at all.  This may have been particularly problematic when strong polarity was present in those sentences.  An example of this is shown in Figure \ref{fig:Snippet1}, where the highlighted portion of the review indicates the sarcastic segment submitted by the annotator, and \textit{awesome}, the most polar word in the entire review (circled), is outside that highlighted sentence.  

Since tweets are self-labeled, users' own varying definitions of sarcasm lead to some extreme idiosyncrasies in the kinds of tweets labeled as sarcastic.  Sarcastic tweets were also often dependent upon outside context.  Some examples include (\#sarcasm tags were removed in the actual dataset): \textit{``My daughter's 5th grade play went over as professional, flawless, and well rehearsed as a Trump speech. \#sarcasm,''} \textit{``\#MilanAlessandria Mario Balotelli scored the fifth goal in the 5-0 win.  He should play for the \#Azzurri at \#EURO2016. \#sarcasm,''} and \textit{``Good morning \#sarcasm.''}

Finally, some past research has found that it is more difficult to discriminate between sarcastic and non-sarcastic texts when the non-sarcastic texts contain sentiment \cite{Gonzalez-Ibanez:2011:IST:2002736.2002850,ghosh-guo-muresan:2015:EMNLP}.  Since our non-sarcastic tweets are emotionally-charged, our classifier may have exhibited lower performance than it would have with only neutral non-sarcastic tweets.  Since distinguishing between literal and sarcastic sentiment is useful for real-world applications of sarcasm detection, we consider the presence of sentiment in our dataset to be a worthwhile challenge.

Regarding the general features developed for this work, the polarity- and subjectivity-based features performed well, while performance using only PMI features was lower.  PMI scores in particular may have been negatively impacted by common Twitter characteristics, such as the trend to join keywords together in hashtags, and the use of acronyms that are unconventional in other domains.  These issues could be addressed to some extent in the future via word segmentation tools, spell-checkers, and acronym expansion.

\section{Conclusions}

This work develops a set of domain-independent features and demonstrates their usefulness for general sarcasm detection.  Moreover, it shows that by applying a domain adaptation step to the extracted features, even a surplus of ``bad'' training data can be used to improve the performance of the classifier on target domain data, reducing error by 14\% relative to prior work.  The Twitter corpus described in this paper is publicly available for research purposes,\footnotemark[2] and represents a substantial contribution to multiple NLP sub-communities.  This shared corpus of tweets annotated for sarcasm will hasten the advancement of further research.  In the future, we plan to extend our approach to detect sarcasm in a completely novel domain, literature, eventually integrating the work into an application to support reading comprehension.

\section*{Acknowledgments}
This material is based upon work supported by the NSF Graduate Research Fellowship Program under Grant 1144248, and the NSF under Grant 1262860.  Any opinions, findings, and conclusions or recommendations expressed in this material are those of the author(s) and do not necessarily reflect the views of the National Science Foundation.

\bibliography{flairs_parde}
\bibliographystyle{aaai}

\end{document}